\documentclass[runningheads]{llncs}
\usepackage{booktabs} % For better looking tables
\usepackage{siunitx} % Allows for alignment at the decimal point
\usepackage{graphicx}
\usepackage{subcaption}
\usepackage[T1]{fontenc}
\usepackage{graphicx}
\usepackage{amsfonts, amsmath}
\usepackage{hyperref}

\begin{document}
%
% \title{Contribution Title\thanks{Supported by organization x.}}
\title{Zero-Shot Underwater Gesture Recognition}
\titlerunning{ZSUGR}
% \author{Sandipan Sarma\inst{1}\orcidID{0000-1111-2222-3333} \and
% Gundameedi Sai Ram Mohan\inst{2,3}\orcidID{1111-2222-3333-4444} \and
% Hariansh Sehgal\inst{3}\orcidID{2222--3333-4444-5555} \and
% Arijit Sur
% }
\author{Sandipan Sarma \and
Gundameedi Sai Ram Mohan \and \\
Hariansh Sehgal \and
Arijit Sur
}
\authorrunning{S. Sarma et al.}
% \institute{Indian Institute of Technology Guwahati, Assam, India \and
\institute{Indian Institute of Technology Guwahati, Assam, India }
\maketitle              % typeset the header of the contribution
\begin{abstract}
Hand gesture recognition allows humans to interact with machines non-verbally, which has a huge application in underwater exploration using autonomous underwater vehicles. Recently, a new gesture-based language called CADDIAN has been devised for divers, and supervised learning methods have been applied to recognize the gestures with high accuracy. However, such methods fail when they encounter unseen gestures in real time. In this work, we advocate the need for zero-shot underwater gesture recognition (ZSUGR), where the objective is to train a model with visual samples of gestures from a few ``seen'' classes only and transfer the gained knowledge at test time to recognize semantically-similar unseen gesture classes as well. After discussing the problem and dataset-specific challenges, we propose new seen-unseen splits for gesture classes in CADDY dataset. Then, we present a two-stage framework, where a novel transformer learns strong visual gesture cues and feeds them to a conditional generative adversarial network that learns to mimic feature distribution. We use the trained generator as a feature synthesizer for unseen classes, enabling zero-shot learning. Extensive experiments demonstrate that our method outperforms the existing zero-shot techniques. We conclude by providing useful insights into our framework and suggesting directions for future research. The code is available at: \url{https://github.com/sandipan211/ZSUGR}.

\end{abstract}
\section{Introduction}
\label{sec:intro}
Despite our planet being 70\% water, our knowledge regarding underwater ecosystems remains limited. A major reason behind this is the harsh underwater environment and human divers face trouble collecting data from ocean depths due to several concerns like increasing hydrostatic pressure and oxygen depletion. In the past few decades, humans have passed the baton to autonomous underwater vehicles (AUVs) that can assist marine experts in several ways by capturing underwater images/videos, detecting oil spillage, inspecting oil and natural gas pipelines, and conducting bathymetric surveys -- all while being well-resistant to the problems faced by human divers. As a result, underwater exploration has garnered a lot of attention nowadays, with applications in oceanography,
marine warfare, information navigation, and marine scene understanding.

\begin{figure}[t]
    \begin{subfigure}{0.49\linewidth}
        \centering
        \includegraphics[width=\linewidth]{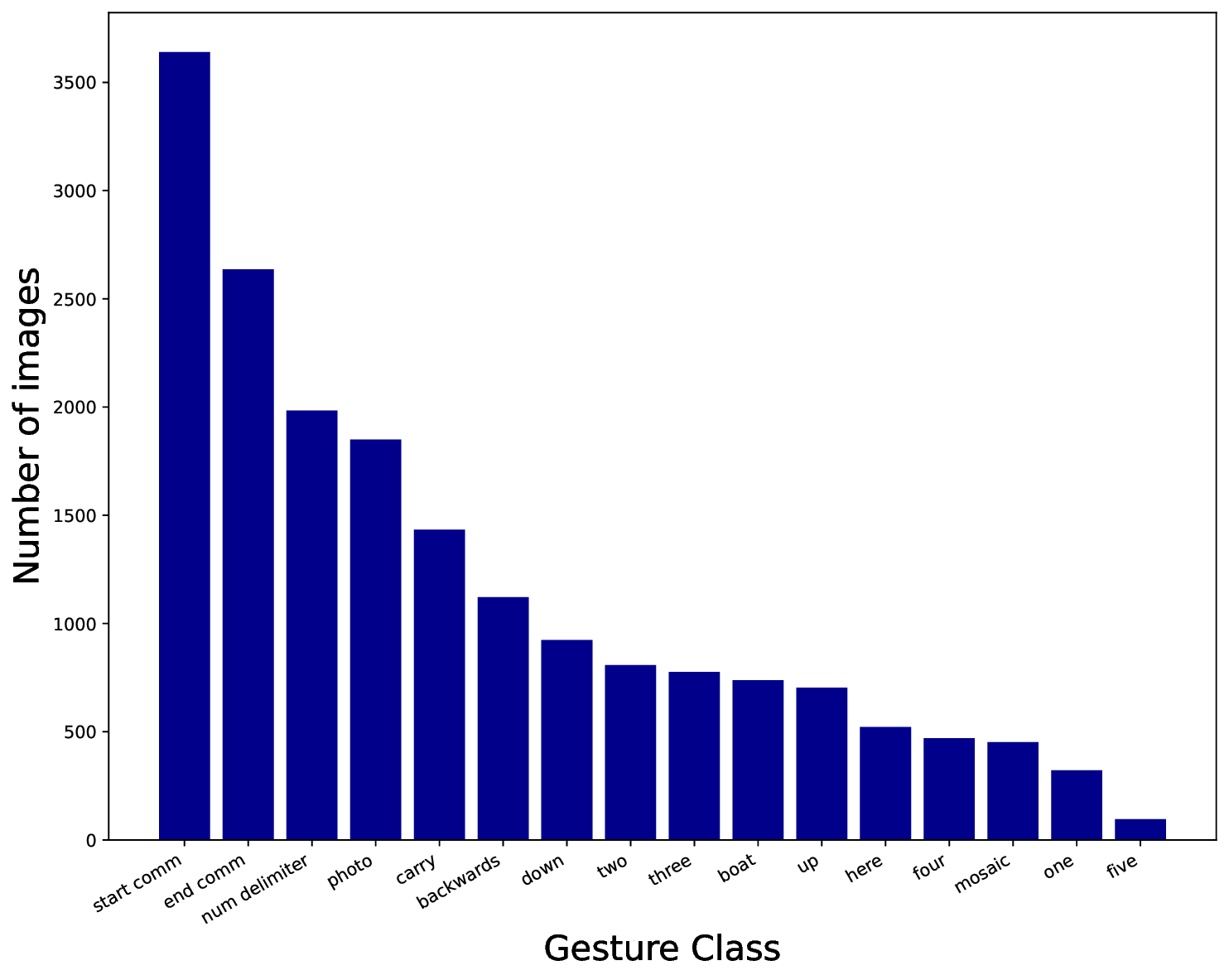}
        \caption{Distribution of gesture classes}
        \label{fig:num_samples}
    \end{subfigure}
    \hfill
    \begin{subfigure}{0.49\linewidth}
        \centering
        \includegraphics[width=0.8\linewidth]{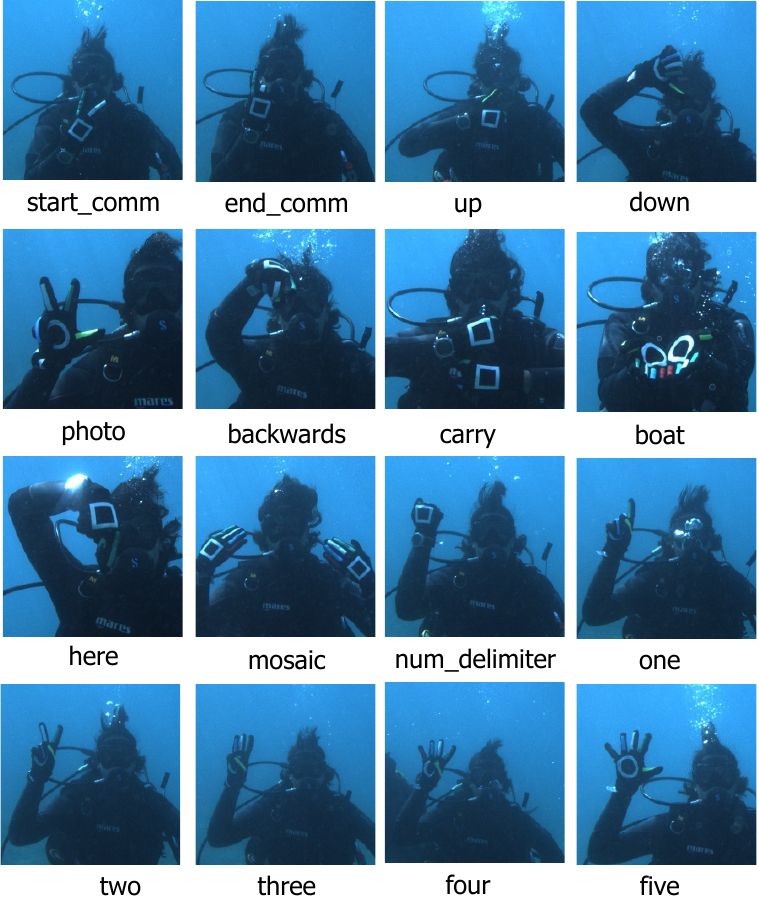}
        \caption{Examples of the 16 gestures}
        \label{fig:gestures}
    \end{subfigure}
    \caption{Properties of the CADDY dataset~\cite{gomez2019caddy}.} 
    \label{fig:caddy}
\end{figure}

% Computer vision plays a vital role in such explorations since AUVs rely on high-quality images to fulfill their objectives and send the collected data to human experts at terrestrial workstations. 
In many underwater missions, AUVs are accompanied by human divers who communicate non-verbally via different gestures. 
%The AUVs must recognize such gestures in real time to perform effectively. 
However, underwater gesture recognition is a relatively underexplored area in computer vision owing to the lack of annotated datasets. Recently, a new gesture-based communication language called CADDIAN~\cite{chiarella2018novel} was developed, and an image-based underwater gesture recognition dataset named CADDY~\cite{gomez2019caddy} (Fig.~\ref{fig:gestures}) was constructed as an effort to facilitate vision research. %like object classification, segmentation, and human pose estimation tasks. 
However, there are primarily two challenges. Firstly, underwater images suffer from problems like low contrast, haziness, color distortion, and blurriness. As a result, traditional gesture recognition methods face problems in analyzing them. Secondly, existing gesture recognition models are predominantly supervised~\cite{yang2019diver,martija2020underwater,chavez2021underwater,zhang2022underwater,yang2023dare,mangalvedhekar2023underwater} and can only recognize gestures from a predefined set used to train the models,  failing to interpret ``unseen'' gestures. This is an inevitable bottleneck as it is impossible to collect thousands of labeled images for every possible gesture that might be used by human divers in the wild. For instance, a new gesture meant to indicate low oxygen levels due to a real-time accident might not be interpreted by an AUV equipped with a supervised gesture recognition model, putting the human diver at risk.

Zero-shot learning (ZSL) has come to the limelight in recent years to alleviate such data scarcity, which mimics the human tendency to learn from other modalities (\textit{semantics} or class attributes) like text or audio in the absence of visual examples. The objective of ZSL is to transfer knowledge about a few ``seen'' concepts/classes via a visual-semantic association and recognize unseen ones. Although a few recent works explore zero-shot gesture recognition, it is still unexplored in underwater scenarios. 

In this paper, we introduce and study the task of \textit{zero-shot underwater gesture recognition} (ZSUGR). Being the first work of its kind, standard seen-unseen zero-shot splits are currently unavailable. Therefore, we propose three new splits by designating seen and unseen classes randomly, following previous splitting strategies~\cite{xu2017transductive,madapana2020zero}. Considering the challenges in extracting strong visual features from underwater images and the highly class-imbalanced nature of the CADDY dataset (Fig.~\ref{fig:num_samples}), a two-stage framework is devised consisting of a novel transformer and a generative adversarial network. In the first stage, we design a novel Gated Cross-Attention Transformer (GCAT) which is responsible for a strong representation learning. 
%We experimentally prove that CNN-based feature extractors like ResNet-101 yield visual features that are largely inadequate in comprehending underwater visual cues. As a resolution, 
 Visual features extracted from a pretrained ResNet-50 are passed to an encoder, and its outputs are transformed into powerful gesture representations by our novel gesture decoder. The trained GCAT is then used as a visual feature extractor for the seen classes. A conditional Wasserstein GAN~\cite{arjovsky2017wasserstein} is then trained with these visual features, with seen class semantic vectors as class conditional variables. Here, we rely on the text encoder of a pretrained visual-language model called CLIP~\cite{radford2021learning} to obtain the semantic vector of a gesture class. Visual features of unseen gestures can then be synthesized using our trained WGAN, enabling us to train a gesture classifier with data from both seen and unseen classes, mitigating the bias problem in zero-shot settings~\cite{paul2019semantically}. 
 %We conduct several experiments and provide quantitative and qualitative results, discussing novel directions for the progress of ZSUGR as a research area. 
 Extensive experiments are conducted, showing the failure of supervised models, our model's improved performance compared to state-of-the-art zero-shot classification methods in conventional and generalized zero-shot settings (inspired by previous zero-shot gesture recognition works~\cite{madapana2020zero,madapana2018hard,wu2021prototype}), and demonstrating the role of improved visual representations for ZSUGR. To summarize, we make the following contributions in this paper:
\begin{itemize}
    \item We introduce the problem of zero-shot \textit{underwater} gesture recognition from images for the first time and discuss its real-world applications.
    \item Seen-unseen splits of gesture classes are proposed for the CADDY dataset for zero-shot training and model evaluation.
    \item A two-stage network is proposed for ZSUGR consisting of a novel transformer that can be trained as a strong underwater visual feature extractor. The obtained features can be fed to a class-conditional generative adversarial network that can learn to synthesize visual features of unseen gestures.
    
\end{itemize}

\section{Related Works}
\label{sec:related}
\subsection{Hand gesture recognition (HGR)}
\label{sec:hgr}
 
 %Gesture cues have been catering to the needs of the hearing and speech-impaired for decades and have now found numerous applications in human-computer interactions. HGR is an area of enormous research interest, specifically because hands are deformable objects and exhibit different textures and skin colors that can be, at times, difficult to recognize due to background clutter. 
 Vision-based HGR methods
 %are more cost-effective than sensor-based and 
 can be broadly categorized into dynamic (limb movements indicate gesture) and static (hand shape and finger orientations indicate gesture). Among these, segmenting the hand using skin color models to extract visual gesture features is a common practice~\cite{zhou2004static,sun2018research}. With the availability of neural networks pretrained on large-scale datasets, transfer learning approaches~\cite{wu2021research} have also been successful. However, the extracted visual features are affected by factors such as lighting conditions and background clutter. To enable HGR frameworks to recognize a wider range of gestures, 3D methods are explored, where several sensors and cameras are used to obtain temporal volumes~\cite{liang2013gesture} or skeletal joints~\cite{devineau2018deep} to analyze the hand shapes.

\subsection{Zero-shot learning (ZSL)}
\label{sec:zsl}
Image classification remains the most researched vision task exploiting ZSL. The first seen-unseen splits were formally provided for five datasets in the seminal work of Xian et al.~\cite{xian2018zero}, who also popularized the concept of \textit{conventional} (only unseen class samples at test time) and \textit{generalized} (samples from both seen and unseen classes at test time) zero-shot learning settings.
%A wide array of zero-shot classifiers exist today, spanning different ideas such as learning intermediate attributes~\cite{lampert2013attribute,rohrbach2010helps}, compatibility learning~\cite{frome2013devise,akata2015evaluation,akata2015label,xian2016latent,romera2015embarrassingly,kodirov2017semantic}, hybrid learning~\cite{norouzi2013zero,changpinyo2016synthesized}, graph-based methods~\cite{yi2022exploring,wang2021zero}, generative methods~\cite{xian2018feature,narayan2020latent,chen2021free,chen2023deconstructed}, and others~\cite{skorokhodov2020class}.
Meanwhile, gesture recognition as a vision task is less explored. One of the earliest works in zero-shot gesture recognition (ZSGR) unified coordinated natural language, gesture, and context~\cite{thomason2017recognizing} to facilitate human-robot interactions. However, ZSGR was standardized in the work of Madapana and Wachs~\cite{madapana2017zsgl}, who used gestures from two gesture-based datasets to define a total of 13 seen and 8 unseen classes. Furthermore, they provided a list of 13 high-level semantic descriptors to characterize each gesture class. Their zero-shot classifiers were inspired by previous image classification methods~\cite{lampert2013attribute,romera2015embarrassingly}. In an extended work, they further introduced hard zero-shot gesture recognition~\cite{madapana2018hard} where there are only a few visual samples for the seen classes as well. Recently in another work, they introduced a gesture attribute dataset~\cite{madapana2019database} with segmented skeletal data and proposed a joint semantic encoder optimizing reconstruction, semantic, and classification losses~\cite{madapana2020zero}. Another novel dataset was built by Wu et al.~\cite{wu2021research} with 16 seen and 9 unseen classes, and then they proposed an end-to-end prototype learning framework~\cite{wu2021prototype}.

\subsection{Underwater diver gesture recognition}
\label{sec:ugr}
Since the arrival of the CADDY dataset~\cite{gomez2019caddy}, several works have been undertaken in the last four years in underwater gesture recognition (UGR). The first such work~\cite{yang2019diver} used pretrained convolutional networks like ResNet, GoogleNet, and others for transfer learning. Several other works that followed~\cite{martija2020underwater,chavez2021underwater,mangalvedhekar2023underwater} compared the performance of classical and deep learning methods for UGR. On the other hand, DARE~\cite{yang2023dare} demonstrated a hierarchical tree-structured classification scheme. Recently, VT-UHGR~\cite{zhang2022underwater} used a pretrained ViT as the visual encoder and a pretrained BERT as the text encoder for multimodal underwater gesture feature learning. Data augmentation using generative methods has also been used to improve UGR performance. However, most of these methods use transfer learning and do not provide any novel architecture for UGR. Additionally, none of them are zero-shot learners. This paper presents the first work that addresses both of these concerns.

\section{Methodology}
\label{sec:methodology}

\subsection{Problem definition}
\label{sec:prob}
Let we have a training set $\mathcal{D}_{train} = \{ (x_i,y_i)|  \; x_i \in \mathcal{X}_{seen}, y_i \in \mathcal{S} \}$, where $x_i$ is an image from a seen class $y_i$. A separate set of unseen data $\mathcal{D}_{novel} = \{ (x_j,y_j)| \; x_j \in \mathcal{X}_{unseen}, y_j \in \mathcal{U} \}$ is given such that the sets of seen and unseen classes are disjoint, i.e., $\mathcal{S} \; \cap \; \mathcal{U} = \phi$. Additionally, we have a semantic vector $a_i \in \mathbb{R}^{512}$ for each gesture class $y \in \mathcal{S} \, \cup \, \mathcal{U}$. We work in the more realistic inductive zero-shot setting~\cite{xian2018zero} -- where visual examples of unseen classes are unavailable during training -- instead of the transductive setting. Then, the task in conventional zero-shot learning is to learn a classifier $f_{zsl}: \mathcal{X}_{unseen} \; \rightarrow \; \mathcal{U}$. For generalized zero-shot learning, a small subset of $\mathcal{X}_{seen}$ ($\mathcal{X}_{seen}^{sub}$) is extracted as the set of seen samples at test time, following the suggestions in the benchmark ZSL paper~\cite{xian2018zero}. The objective then changes to learning a classifier $  f_{gzsl}: \mathcal{X}_{seen}^{sub} \; \cup \; \mathcal{X}_{unseen} \rightarrow \mathcal{S} \; \cup \; \mathcal{U}$.

\begin{figure}[t]
\centering
\includegraphics[width=\textwidth]{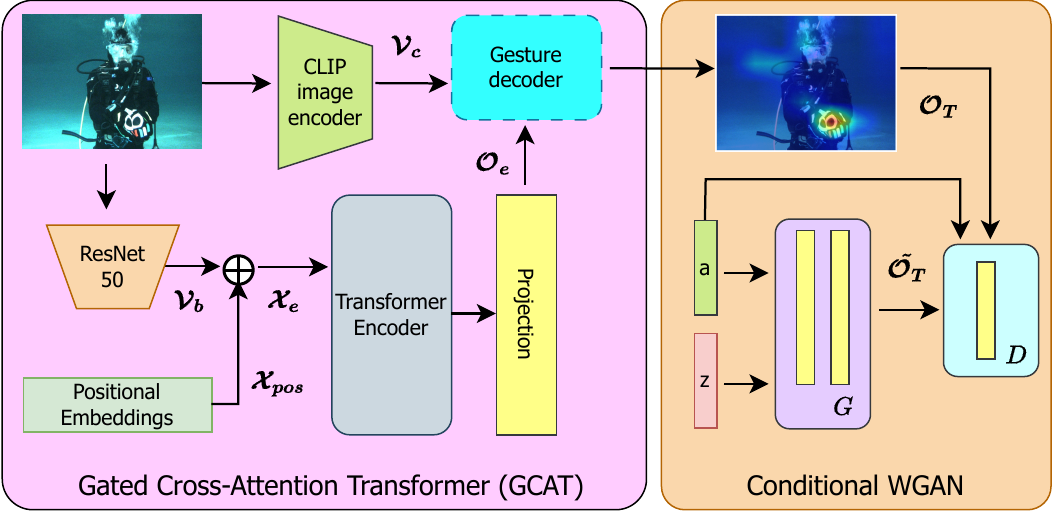}
\caption{Proposed two-stage framework for ZSUGR. Here, $z$ denotes a random noise vector, and $a$ denotes semantic vector of a gesture class.}
\label{fig:arch}
\end{figure}

\subsection{System overview}
\label{sec:overview}
Figure~\ref{fig:arch} shows our overall two-stage framework, which includes a novel transformer and a Wasserstein GAN (WGAN). Given an underwater image $\mathcal{X} \in \mathbb{R}^{C \times H \times W}$, its visual feature map $\boldsymbol{\mathcal{V}_b} \in \mathbb{R}^{C' \times H' \times W'}$ is first extracted using a pretrained ResNet-50 backbone. Since a transformer encoder considers each spatial location in $\boldsymbol{\mathcal{V}_b}$ as a token while computing self-attention, it is necessary to keep track of the token positions. We construct positional embeddings $\boldsymbol{\mathcal{X}_{pos}} \in \mathbb{R}^{C' \times H' \times W'}$, add them element-wise with $\boldsymbol{\mathcal{V}_b}$, and pass the resulting feature map to the transformer encoder $E$. The outputs from $E$ are then decoded via our novel gated cross-attention mechanism guided by powerful visual representations from a pre-trained CLIP. These decoded visual gesture features are used as ``real features'' for training a WGAN, conditioned on class semantics. The trained WGAN can be used to generate visual features corresponding to unseen gesture classes. A classifier $\Phi_{cls}(.)$ is trained with seen class data and the synthesized unseen class data for zero-shot gesture recognition. We discuss the novel GCAT in Sec.~\ref{sec:gcat}, followed by the WGAN in Sec.~\ref{sec:wgan}.

\subsection{Gated cross-attention transformer (GCAT)}
\label{sec:gcat}
\begin{figure}[t]
\centering
\includegraphics[scale=0.5]{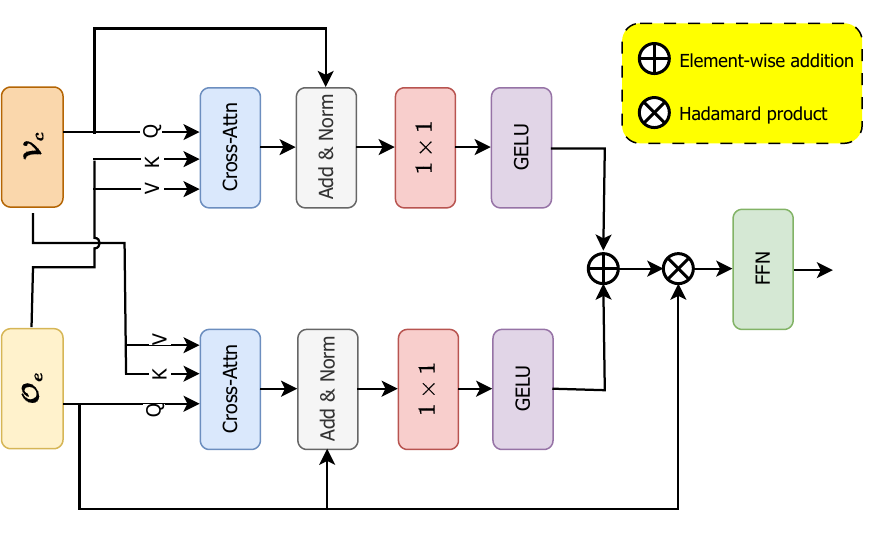}
\caption{Architecture of our novel GCAT gesture decoder.} \label{fig:decoder}
\end{figure}

As mentioned in Sec.~\ref{sec:overview}, the encoder $E$ takes the input:
\begin{equation}
    \label{eq:enc_input}
    \boldsymbol{\mathcal{X}_e} = \boldsymbol{\mathcal{V}_b} + \boldsymbol{\mathcal{X}_{pos}}
\end{equation}
and performs a self-attention operation on the visual tokens. The output $\boldsymbol{\mathcal{O}_e} \in \mathbb{R}^{C' \times H'W'}$ contains image-wide contextual information that is passed to our gesture decoder $Dec$. Meanwhile, we extract additional visual features $\boldsymbol{\mathcal{V}_c} \in \mathbb{R}^{C'' \times k}$ from a pretrained CLIP image encoder and use them to refine the knowledge coming from the encoder, yielding gesture features. To this end, $\boldsymbol{\mathcal{O}_e}$ is first projected to CLIP dimension, and then a two-branch cross-attention mechanism is proposed. The two branches perform identical operations, with the only difference being in the query ($Q$), key ($K$), and value ($V$) assignments. The branch assignments (Fig.~\ref{fig:decoder}) are as follows:
\begin{align}
    \label{eq:qkv}
    Q_L = \boldsymbol{\mathcal{O}_e}, \quad K_L = V_L = \boldsymbol{\mathcal{V}_c} \\
    Q_R = \boldsymbol{\mathcal{V}_c}, \quad K_R = V_R = \boldsymbol{\mathcal{O}_e}
\end{align}
where $L$ and $R$ denote left and right branches. The attention outputs are normalized using layer normalization (LN) operation and added to the branch queries:
\begin{align}
    \label{eq:addnorm}
    A_L = LN(Q_L + CrossAtt(Q_L, K_L, V_L)) \\
    A_R = LN(Q_R + CrossAtt(Q_R, K_R, V_R))
\end{align}

Now, since visual inputs to $Dec$ come from two sources -- $\boldsymbol{\mathcal{O}_e}$ based on ResNet features and $\boldsymbol{\mathcal{V}_c}$ based on CLIP features -- it is beneficial to weight the attention outputs for understanding the contribution of the encoder outputs as well as CLIP context. Consequently, we propose a gating mechanism where we apply a $1 \times 1$ convolution first to the attention outputs of each branch and downscale them by a factor of 2. A GELU activation~\cite{hendrycks2016gaussian} finally produces the gated values. The gating function
can hence be formulated as:
\begin{align}
    \label{eq:gate}
    g (A_L) = GELU (\boldsymbol{w_L} A_L + \boldsymbol{b_L}) \\
    g (A_R) = GELU (\boldsymbol{w_R} A_R + \boldsymbol{b_R})
\end{align}
where $(\boldsymbol{w_L}, \boldsymbol{b_L})$ and $(\boldsymbol{w_R}, \boldsymbol{b_R})$ denote the weights and biases of the convolutional layer in the left and right branches respectively. The gated outputs indicate how much cross-attention information should be preserved and how much can be learned subsequently. The attention maps from the two branches are fused together, and the source knowledge from the encoder is refined as:
\begin{align}
    \label{eq:updated_enc}
    \boldsymbol{\mathcal{O}_e} = \boldsymbol{{\mathcal{O}_e}} \circ (A_L + A_R)
\end{align}
where $\circ$ denotes Hadamard product. It is then passed to a feedforward network (FFN), as shown in Fig.~\ref{fig:decoder}, to produce the final feature map $\boldsymbol{\mathcal{O}_T}$ containing gesture cues. A channel-wise mean is taken at the end such that $\boldsymbol{\mathcal{O}_T} \in \mathbb{R}^d$. 

\subsection{Transformed-feature generating network}
\label{sec:wgan}
Due to problems like color distortion and haziness that underwater images suffer from, it becomes exceedingly difficult to work in extreme scenarios like zero-shot in the absence of visual images of unseen classes. Moreover, predictions may get highly biased toward the seen classes owing to the imbalanced nature of underwater datasets like CADDY. To alleviate the seen-bias problem, we employ a conditional Wasserstein Generative Adversarial Network (c-WGAN)~\cite{arjovsky2017wasserstein}. As depicted in Fig.~\ref{fig:arch}, it takes the visual gesture features $\boldsymbol{\mathcal{O}_T}$ corresponding to seen classes as input and their semantic vectors ($a$) as a conditional variable. The generator $G$ attempts to mimic the true distribution of the visual features $\boldsymbol{\mathcal{O}_T}$ while being discriminative in the visual space with guidance received from the class semantics. It generates seen class features $\boldsymbol{\mathcal{\Tilde{O}}_T}$, which are then fed to a critic network $D$ along with features from true distribution $\boldsymbol{\mathcal{O}_T}$. $D$ acts as a binary classifier that scores the ``realness'' of the input data (either coming from the mimicked feature distribution or from the true distribution). During backpropagation, $D$ tries to improve its predictions, whereas $G$ tries to improve itself based on the feedback from $D$ and synthesizes features closer to the true distribution. For optimization, we use the Wasserstein loss:
\begin{align}
    \label{eq:wgan}
    \mathcal{L}_{\tt WGAN} = \mathbb{E}[D(\boldsymbol{\mathcal{O}_T}, a)] - \mathbb{E}[D(\Tilde{\boldsymbol{\mathcal{O}_T}}, a)] + \lambda \mathbb{E}[(|| \nabla_{\boldsymbol{\hat{\mathcal{O}_T}}} D(\hat{\boldsymbol{\mathcal{O}_T}}, a) ||_2 -1)^2]
\end{align}

where the first two terms represent the critic loss in WGAN and the third term represents a gradient penalty~\cite{gulrajani2017improved}, with $\lambda$ being the penalty coefficient. $\Tilde{\boldsymbol{\mathcal{O}_T}} = G(z, a)$ denotes the generated feature, and $\boldsymbol{\hat{\mathcal{O}_T}} = \rho \, \boldsymbol{\mathcal{O}_T} + (1 - \rho) \, \Tilde{\boldsymbol{\mathcal{O}_T}}$, with $\rho \sim U(0,1)$. Additionally, we use a mode-seeking loss~\cite{mao2019mode} to mitigate the problem of mode-collapse in GANs:
\begin{align}
    \label{eq:mode}
    \mathcal{L}_{\tt MS} = \mathbb{E}[ || G(z_1, a) - G(z_2, a) ||_1 / || z_1 - z_2 ||_1]
\end{align}
Here, $z_1$ and $z_2$ correspond to two noise vectors that produce different visual features. The idea is to regularize that the generated features should get sampled corresponding to noise vectors in the learned distribution that are far apart from each other, giving rise to discriminative synthetic visual features.
\begin{table}[t]
\centering
\caption{Details of proposed seen-unseen splits of the CADDY dataset.}
\label{tab:dataset_splits}
\begin{tabular}{c @{\hspace{1em}} c @{\hspace{1em}} c @{\hspace{1em}} c @{\hspace{1em}} c @{\hspace{1em}} c}
\toprule
% \rowcolor{lightgray}
Split & Seen classes & Unseen classes & $\mathcal{X}_{seen}$ &  $\mathcal{X}_{seen}^{sub}$ & $\mathcal{X}_{unseen}$ \\
\midrule
1 & 10 & 6 & 12505 &  1389 & 4584 \\
2 & 10 & 6 & 10251 & 1139 & 7088 \\
3 & 10 & 6 & 11713 & 1301 & 5464 \\
\bottomrule
\end{tabular}
\end{table}

\subsection{Training and inference}
\label{sec:train-test}
We follow a two-stage training process, as shown in Fig.~\ref{fig:arch}. In the first stage, we train the transformer GCAT to produce visual gesture features. We feed them to a classifier $\Phi_{c}$ whose weights are initialized by the text embeddings (or semantics) of the seen classes. For a class $y_i$, we obtain its semantic vector by converting the class name to ``A photo of a diver gesturing [$y_i$]'' via a prompt template $\rho_g(.)$, and then using a pretrained CLIP text encoder~\cite{radford2021learning} $\psi(.)$ as: 
\begin{align}
    \label{eq:semantics}
\boldsymbol{a_i} = \psi(\rho_g(y_i)) \quad \forall \,y_i \in \mathcal{S} \cup \mathcal{U}
\end{align}
We minimize a cross-entropy loss ($\mathcal{L}_{\tt CE}$) between the predicted and ground-truth gesture labels, strengthening the visual feature representations for seen gestures.

The trained GCAT is used in the next stage as a feature extractor for underwater images. With the extracted features (seen data), the c-WGAN is trained to synthesize seen data by minimizing the loss (with hyperparameter $\alpha$):
\begin{align}
    \label{eq:gan_loss}
    \underset{G}{min} \; \underset{D}{max} \quad  \mathcal{L}_{\tt WGAN} + \alpha \,\mathcal{L}_{\tt MS}
\end{align} 
Given unseen class semantic vectors, the trained GAN can then be used to generate unseen class visual features. The synthesized unseen features and the visual features for seen classes extracted from the trained GCAT can then be combined and used to train a linear softmax classifier $\Phi_{cls}(.)$ for zero-shot prediction. At test time, visual gesture features of an image are first extracted using GCAT, and the trained classifier $\Phi_{cls}(.)$ predicts the gesture class. 
\begin{table}[t]
\centering
\caption{Top-1 accuracy (in \%) reported for CZSL and GZSL settings. H denotes harmonic mean of seen and unseen accuracy in GZSL. Best results are in \textbf{bold}. Codes for CZSL training and evaluation were not provided by  DGZ~\cite{chen2023deconstructed}.}
\label{tab:existing_zsl}
% \begin{tabular}{lcccc}
\begin{tabular}{l@{\hspace{10pt}} |  c@{\hspace{10pt}} | c@{\hspace{10pt}} c@{\hspace{10pt}} c}

\toprule
\textbf{Method} & $\boldsymbol{U_{czsl}}$ & $\boldsymbol{S_{gzsl}}$ & $\boldsymbol{U_{gzsl}}$ & $\boldsymbol{H}$ \\
\midrule
TFVAEGAN~\cite{narayan2020latent}    & $41.50 \pm 5.47$ & $79.57 \pm 13.11$ & $13.49 \pm 5.28$ & $22.51 \pm 7.40$ \\
CNZSL~\cite{skorokhodov2020class} & $16.72 \pm 0.08$ & $20.27 \pm 3.84$ & $11.88 \pm 2.78$ & $14.97 \pm 3.23$ \\
FREE~\cite{chen2021free}     & $15.83 \pm 3.95$ & $84.88 \pm 9.03$ & $14.55 \pm 3.36$ & $24.61 \pm 4.50$ \\
CE-GZSL~\cite{han2021contrastive} & $39.89 \pm 6.49$ & $\boldsymbol{94.11 \pm 0.55}$ & $2.58 \pm 0.45$ & $5.01 \pm 0.87$ \\
DGZ~\cite{chen2023deconstructed}    & - & $57.89 \pm 2.90$ & $15.72 \pm 3.32$ & $24.62 \pm 4.14$ \\

\midrule
\textbf{Ours}  & $\boldsymbol{45.91 \pm 4.71}$ & $61.93 \pm 5.71$ & $\boldsymbol{20.03 \pm 7.14}$ & $\boldsymbol{29.53 \pm 7.06}$ \\
\bottomrule
\end{tabular}
\end{table}

\section{Experiments}
\label{sec:expt}

\subsection{Dataset}
\label{sec:dataset}
All our experiments are conducted on the CADDY underwater stereo-vision dataset~\cite{gomez2019caddy}, one of the largest publicly available underwater image datasets and the only one for diver gesture recognition. 
%The images are collected from different locations, such as open sea and indoor and outdoor pools -- exhibiting a wide range of image qualities, entropy density distributions, contrast, etc.
% For improved visual gesture feature extraction, the diver gloves were modified by adding color stripes to the fingers so that they are distinguishable from the divers' black body suit. 
Overall, 18,478 labeled diver gesture images are present in CADDY, belonging to 16 different gesture classes (Fig.~\ref{fig:gestures}). The gestures used by divers correspond to the CADDIAN sign language~\cite{chiarella2018novel}. Notably, the dataset is highly imbalanced with a long-tail distribution (Fig.~\ref{fig:num_samples}), making it very challenging for zero-shot learning.  

\subsection{Zero-shot splits and evaluation protocols}
\label{sec:splits-eval}
Due to the absence of any previous work on ZSUGR, we define new splits of the CADDY dataset, marking the classes to be used as seen/unseen for zero-shot models. In previously undertaken zero-shot vision applications, seen-unseen splits are proposed following two approaches. Some works designate fixed sets of seen/unseen classes~\cite{xian2018zero} for a given dataset. On the other hand, some works randomly split the dataset classes into seen and unseen sets at a 50\%/50\% proportion, and results are reported over multiple such random splits~\cite{xu2017transductive}. We take this second approach to eliminate human bias in picking the seen-unseen split. However, owing to highly imbalanced data, 
%a 50\%/50\% proportion does not give us adequate number of images to train a zero-shot classifier. Hence, 
we propose to pick 10 seen and 6 unseen classes randomly from the dataset, and three such splits are obtained. Table~\ref{tab:dataset_splits} provides our split details. 

For model evaluation, we follow the benchmark protocol for zero-shot classification~\cite{xian2018zero}. For conventional zero-shot (\textbf{CZSL}) evaluation, we report the top-1 accuracy for the unseen classes ($\boldsymbol{U_{czsl}}$). In the case of generalized evaluation (\textbf{GZSL}), we compute the top-1 accuracy for both seen ($\boldsymbol{S_{gzsl}}$) and unseen ($\boldsymbol{U_{gzsl}}$) classes and report their harmonic mean ($\boldsymbol{H}$). The harmonic mean encourages the model to perform well on both seen and unseen classes, which is a more realistic metric for zero-shot settings. Since we have multiple random splits, we indicate the mean and standard deviation of obtained accuracy across the three random splits for all the reported results. 

\begin{figure}[t]
    \centering
    \begin{subfigure}{0.48\linewidth}
        \centering
        \includegraphics[width=\linewidth]{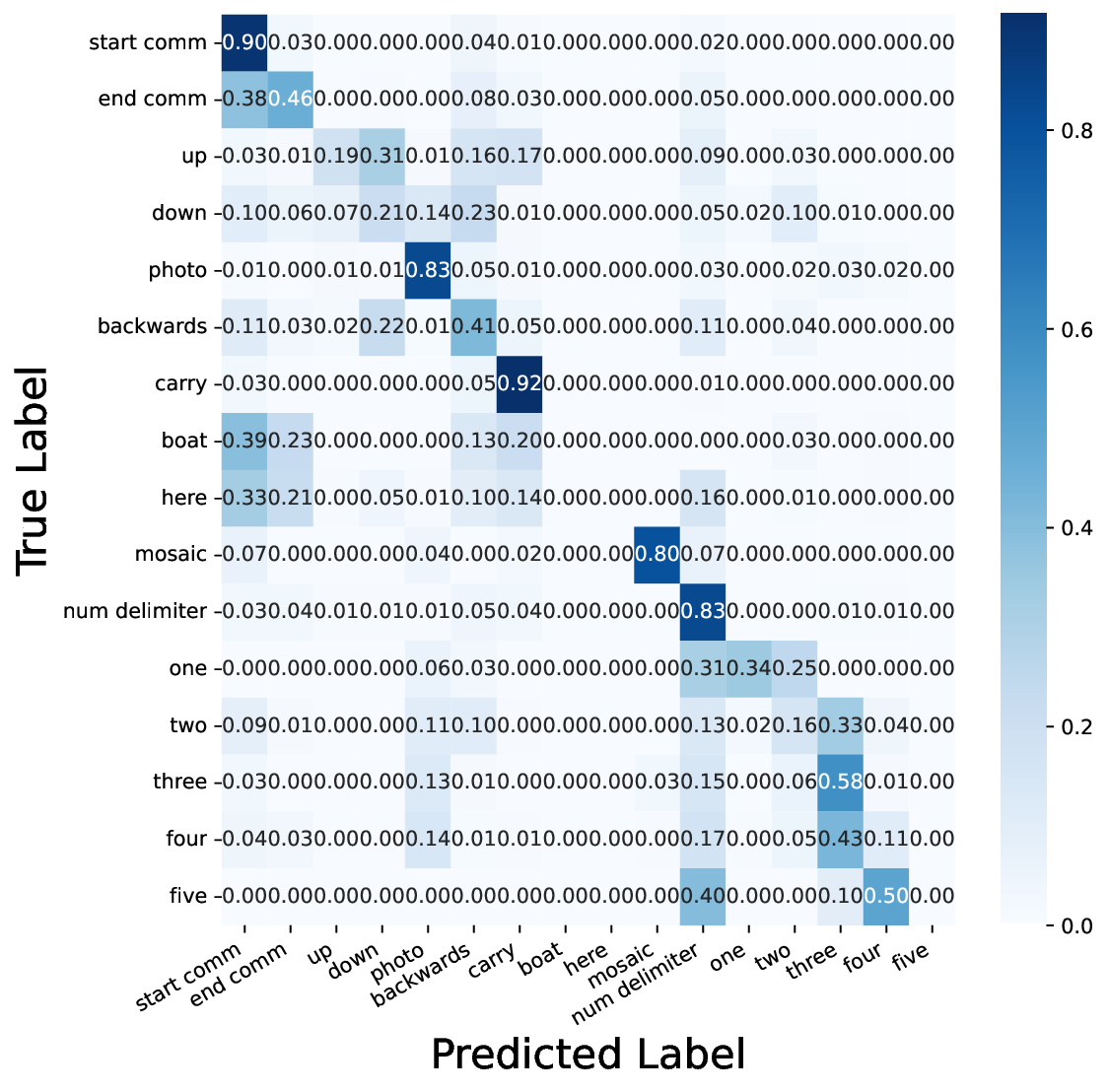}
        \caption{DGZ~\cite{chen2023deconstructed}}
        \label{fig:cm_DGZ}
    \end{subfigure}
    \hfill
    \begin{subfigure}{0.48\linewidth}
        \centering
        \includegraphics[width=\linewidth]{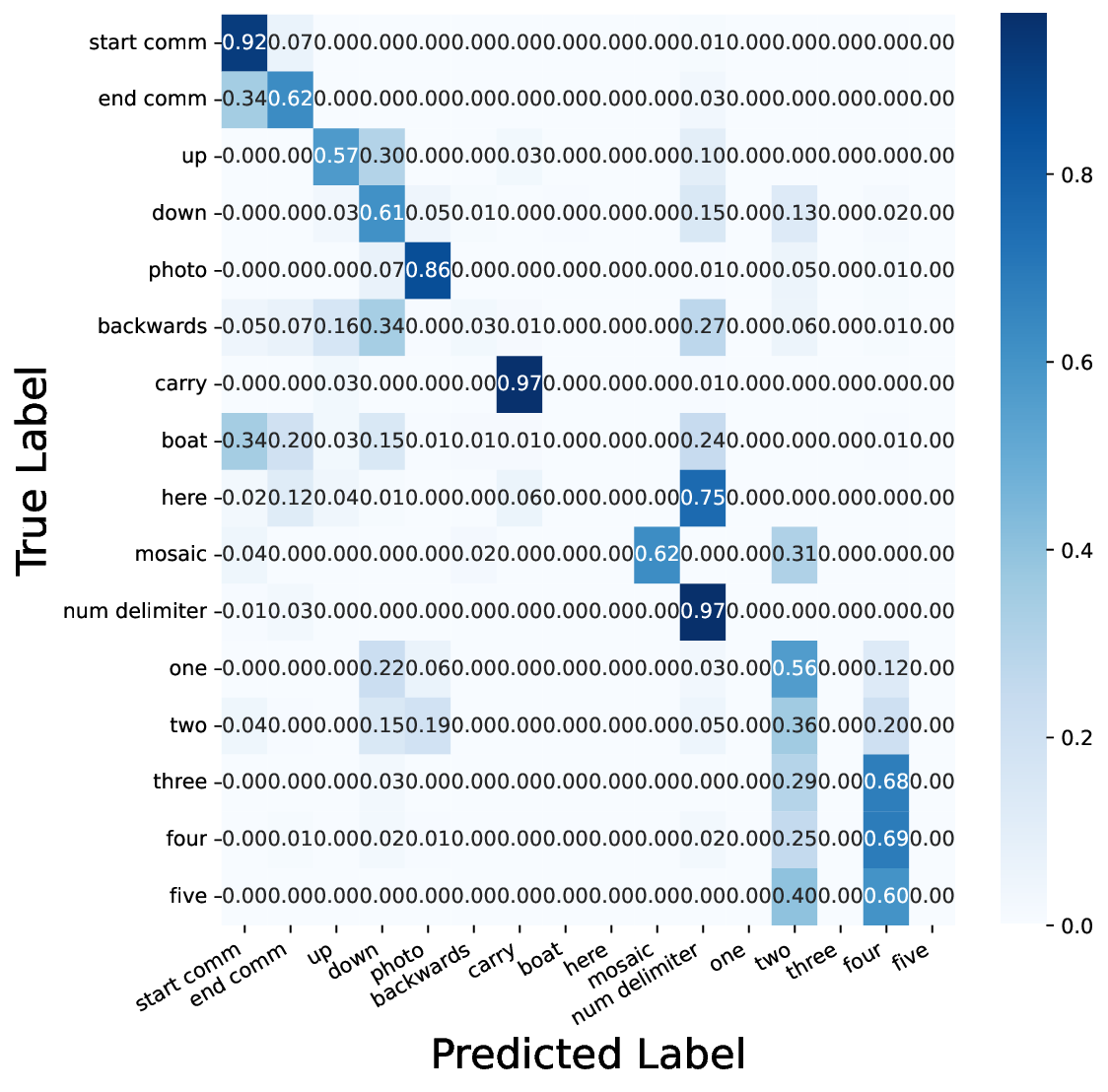}
        \caption{Ours}
        \label{fig:cm_ours}
    \end{subfigure}
    \caption{Comparison of our GZSL confusion matrix with the state-of-the-art.}
    \label{fig:cm}
\end{figure}

\subsection{Implementation details}
\label{sec:imple}
All the experiments are undertaken in PyTorch using a single NVIDIA A100 GPU. The raw underwater images are resized to $3 \times 224 \times 224$. We used the ResNet-50 backbone to extract a feature map $\boldsymbol{\mathcal{V}_b}$ of $256 \times 7 \times 7$. Our encoder $E$ contains the same layers as the traditional transformer, consisting of a self-attention mechanism, followed by layer normalization and neural layers. The final encoder output is obtained after 3 blocks of encoder layers. We use 3 blocks of decoder layers, and the final decoder output $\boldsymbol{\mathcal{O}_T} \in \mathbb{R}^{(7 \times 7) \times 512}$ is averaged across the feature maps to yield a 512-dimensional visual gesture feature. The transformer GCAT is optimized using AdamW with a learning rate of 1e-5 and weight decay of 1e-4. The output from the CLIP image encoder $\boldsymbol{\mathcal{V}_c}$ used for refining the encoder features is a feature map of dimension ${768 \times 50}$. As for the CLIP text encoder, the output embeddings are 512-dimensional vectors used as semantic vectors for every class. In our c-WGAN, the generator ($G$) and critic ($D$) networks are essentially two multi-layer perceptions, both optimized using the Adam optimizer with a learning rate of 1e-4. In $\mathcal{L}_{\tt MS}$, we use $\alpha = 1e-4$.

\begin{figure}[t]
    \centering
    \begin{subfigure}{0.24\linewidth}
        \centering
        \includegraphics[width=\linewidth]{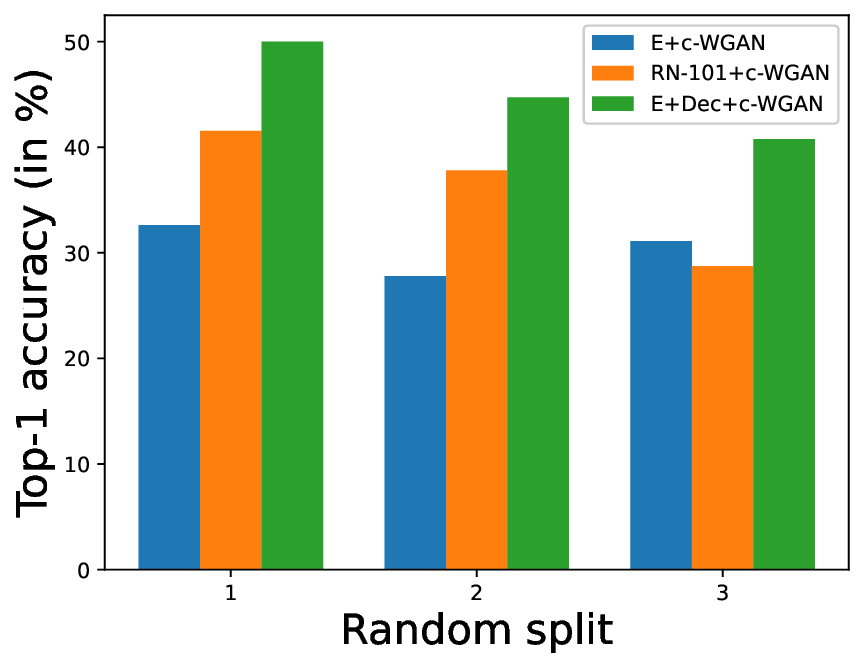}
        \caption{$\boldsymbol{U_{czsl}}$}
        \label{fig:czsl}
    \end{subfigure}
    \hfill
    \begin{subfigure}{0.24\linewidth}
        \centering
        \includegraphics[width=\linewidth]{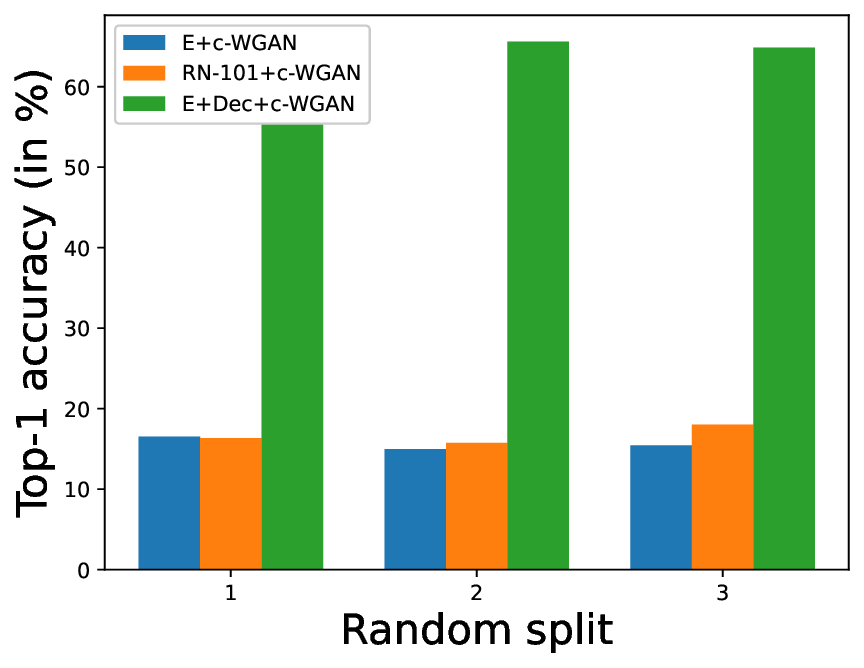}
        \caption{$\boldsymbol{S_{gzsl}}$}
        \label{fig:s-gzsl}
    \end{subfigure}
    \hfill
    \begin{subfigure}{0.24\linewidth}
        \centering
        \includegraphics[width=\linewidth]{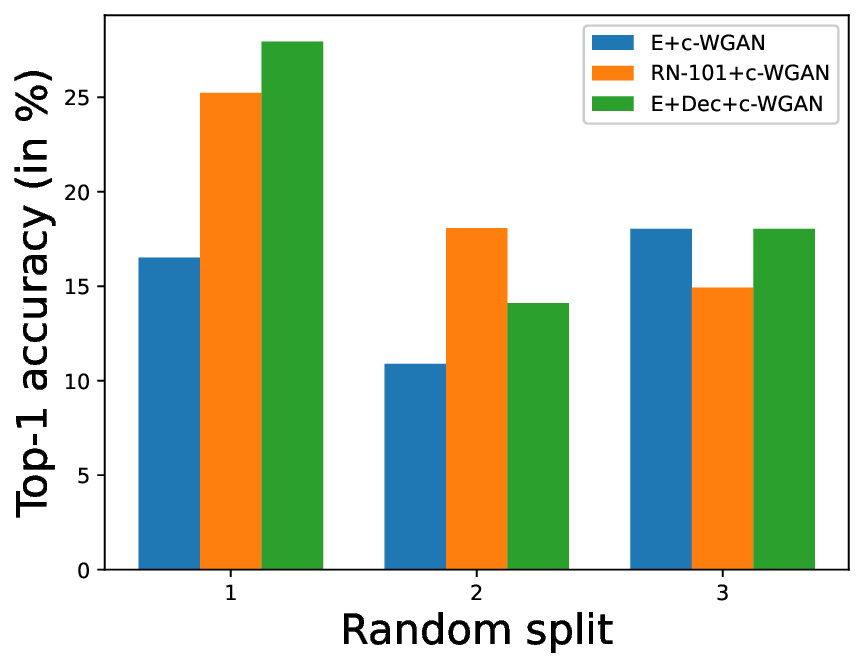}
        \caption{$\boldsymbol{U_{gzsl}}$}
        \label{fig:u-gzsl}
    \end{subfigure}
    \hfill
    \begin{subfigure}{0.24\linewidth}
        \centering
        \includegraphics[width=\linewidth]{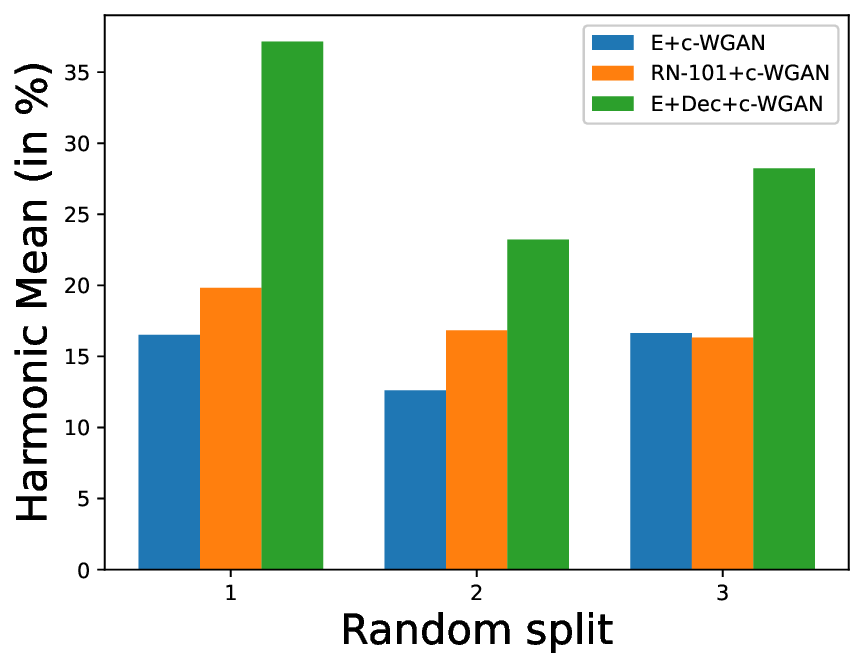}
        % harmonic mean (in %) 
        % green legend E+Dec+WGAN
        \caption{$\boldsymbol{H}$}
        \label{fig:hm}
    \end{subfigure}
    \caption{Component analysis of our framework for the three proposed random splits. Top-1 accuracy is reported in CZSL ($\boldsymbol{U_{czsl}}$) and GZSL settings ($\boldsymbol{S_{gzsl}}$ and $\boldsymbol{U_{gzsl}}$ with harmonic mean $\boldsymbol{H}$). E = GCAT encoder, RN-101 = ResNet-101 as feature extractor, Dec = GCAT decoder, c-WGAN = Conditional WGAN.}
    \label{fig:components}
\end{figure}

\subsection{Zero-shot results}
Since there are no prior works in ZSUGR, we adapted a few zero-shot image classification methods for the ZSUGR task, following previous works~\cite{madapana2018hard,wu2021prototype,madapana2020zero}. For a fair comparison, we train these methods with the same semantic vectors as ours. We evaluate these models in both CZSL and GZSL settings and report their mean and standard deviations over the three random splits we proposed.

\subsubsection{Conventional setting (CZSL).} DGZ~\cite{chen2023deconstructed} did not provide their implementation for CZSL evaluation. CZSL results for some other methods are reported in Tab.~\ref{tab:existing_zsl}. Our method outperforms all of them by a good margin with an average top-1 accuracy of $45.91 \pm 4.71\%$.

\begin{figure}[t]
\centering
\includegraphics[width=0.9\textwidth]{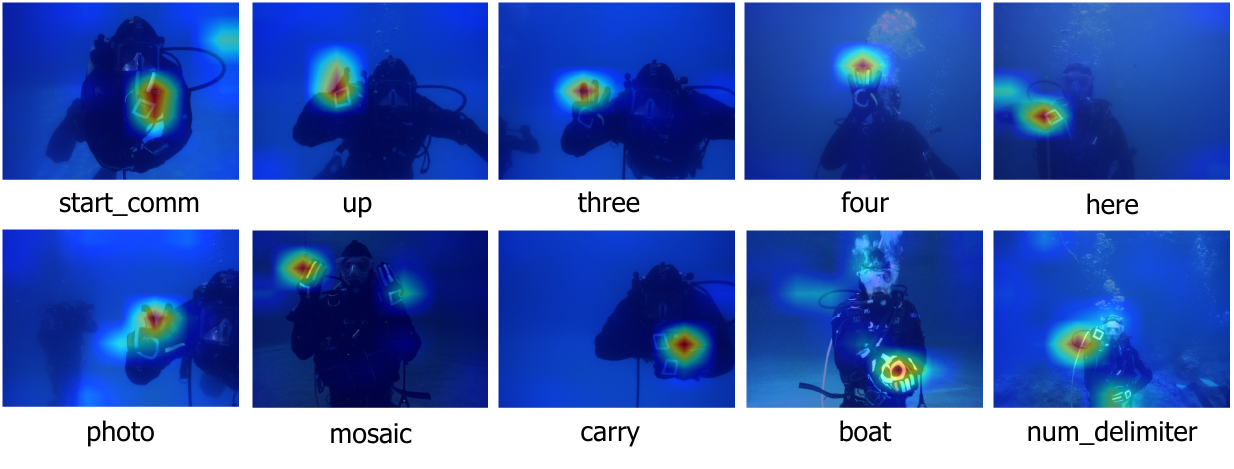}
\caption{Attention visualization of GCAT gesture decoder for ten gestures.} \label{fig:decoder_att}
\end{figure}

\subsubsection{Generalized setting (GZSL).} Table~\ref{tab:existing_zsl} suggests that CE-GZSL~\cite{han2021contrastive} is extremely capable of recognizing seen gestures during evaluation with an accuracy of $94.11 \pm 0.55\%$ but performs very poorly in case of unseen gesture classes with an accuracy of $2.58 \pm 0.45\%$, yielding low harmonic mean. This indicates that CE-GZSL has low generalizability in ZSUGR for the CADDY dataset, although it exhibited state-of-the-art (SOTA) generalizability in image classification. The other methods suffer from the same issue as well. One of the reasons behind such low generalization ability could be the class imbalance problem in CADDY, which is not as extreme in the image classification datasets on which these methods are usually evaluated. Moreover, the zero-shot methods usually suffer from the bias problem~\cite{paul2019semantically}, where predictions for unseen class samples are incorrectly biased towards a similar seen class. Together, the bias problem and class imbalance make ZSUGR a challenging task for the SOTA zero-shot classification methods. Meanwhile, our method seemingly finds a better balance between seen and unseen class knowledge. Additionally, it deals with the bias problem better than the other generative methods like TFVAEGAN~\cite{narayan2020latent}, FREE~\cite{chen2021free}, and DGZ~\cite{chen2023deconstructed}, indicating that our GAN has had superior training with gesture features extracted from the novel GCAT, as compared to the ResNet-101 features with which the GANs of other methods are trained. Consequently, we obtain the best harmonic mean of $29.53 \pm 7.06\%$ (Tab.~\ref{tab:existing_zsl}). For class-wise comparison, we plot confusion matrices in Fig.~\ref{fig:cm} to demonstrate how the SOTA classification method DGZ~\cite{chen2023deconstructed} fares against our method. 
\begin{table}[t]
\centering
\caption{Effect of activation function on top-1 accuracy (in \%) during CZSL and GZSL evaluation. H denotes the harmonic mean of seen and unseen accuracy.}
\label{tab:actFn}
% \begin{tabular}{lcccc}
\begin{tabular}{l@{\hspace{10pt}} | c@{\hspace{10pt}} | c@{\hspace{10pt}} c@{\hspace{10pt}} c}

\toprule
\textbf{Activation} & $\boldsymbol{U_{czsl}}$ & $\boldsymbol{S_{gzsl}}$ & $\boldsymbol{U_{gzsl}}$ & $\boldsymbol{H}$\\
\midrule
GELU    & $45.16 \pm 4.64$  & $61.93 \pm 5.71$ & $20.03 \pm 7.14$ & $29.53 \pm 7.06$ \\
ELU     & $44.66 \pm 8.11$ & $59.50 \pm 12.11$  &$18.05 \pm 5.94$ & $26.84 \pm 6.54$ \\
RELU    & $45.38 \pm 8.74$ & $65.48 \pm 2.03$ & $14.73 \pm 4.17$ & $23.86 \pm 6.65$ \\
Sigmoid & $40.71 \pm 5.25$ & $36.50 \pm 5.52$ & $30.02 \pm 2.81$ & $32.67 \pm 2.15$ \\
SiLU    & $47.71 \pm 3.75$ & $65.65 \pm 5.09$ & $15.46 \pm 0.74$ & $25.03 \pm 1.24$ \\
\bottomrule
\end{tabular}
\end{table}

\subsubsection{Qualitative analysis.} Fig.~\ref{fig:decoder_att} shows decoder attention visualizations highlighting where our GCAT gesture decoder is focusing at test time while extracting visual gesture features. 
%Due to space constraints, we have shown attention visualizations corresponding to samples from ten gesture classes only. 
We observe that our trained GCAT focuses on the hands of the divers, resulting in highly relevant visual features extracted for ZSUGR. Specifically, for relevant gestures like \textit{three}, \textit{photo}, and \textit{four}, the decoder focuses on the fingers, which are more informative. On the other hand, it shifts its focus to both hands whenever relevant, whether the hands are close to each other (like \textit{boat} and \textit{carry}) or distant (such as \textit{mosaic}).

\subsubsection{Plight of supervised models.}
Previous works~\cite{yang2019diver,martija2020underwater,mangalvedhekar2023underwater} have shown that pretrained CNN models can be easily adapted to underwater gesture recognition via transfer learning on CADDY. In Tab.~\ref{tab:supervised}, we mention the supervised gesture recognition results for CADDY as reported by previous works, with the best accuracy reaching 98\% using ResNet-18. We employ these models in the GZSL setting to investigate their performance in recognizing unseen gestures. To this end, we obtain visual features from these pretrained CNNs via transfer learning and compute a cosine similarity with the CLIP semantic vectors to get zero-shot predictions.  It can be observed from Tab.~\ref{tab:supervised} that their performance drops for both seen and unseen classes. Specifically, they fail miserably in recognizing unseen gestures, with a best harmonic mean of $2.45 \pm 4.18\%$ achieved using MobileNet-v3. On the contrary, our model achieves a harmonic mean of $29.53 \pm 7.06\%$. Hence, the need for zero-shot models for underwater gesture recognition is evident as the best-performing supervised models are insufficient.

\subsection{Ablation studies}

\subsubsection{Component analysis}
To validate the effect of each component, we conduct an ablation study on CADDY in both CZSL and GZSL settings with the following changes to our framework: (i) removing the novel GCAT gesture decoder and using only the vanilla encoder to get visual features and (ii) removing the GCAT module and using pretrained ResNet-101 visual features to train our c-WGAN. Fig.~\ref{fig:components} suggests that our decoder significantly improves performance in both CZSL and GZSL settings across all three random splits. This justifies refining the visual features extracted by the vanilla encoder using our novel decoder, which makes GCAT attend to more critical and discriminative regions of the diver's hand within the image. On the contrary, the raw features from ResNet-101 or the vanilla encoder alone are less effective.

\begin{table}[t]
\centering
\caption{A performance comparison of pretrained CNN models in the supervised (done by previous works) and generalized zero-shot settings (shown by us). Top-1 accuracy (in \%) is reported for supervised setting, whereas top-1 accuracy for seen and unseen classes are reported over our proposed splits for GZSL.}
\label{tab:supervised}
% \begin{tabular}{lcccc}
\begin{tabular}{l@{\hspace{10pt}} | c@{\hspace{10pt}} | c@{\hspace{10pt}} c@{\hspace{10pt}} c@{\hspace{10pt}} }

\toprule
\textbf{Pretrained CNN} &\textbf{Supervised} & $\boldsymbol{S_{gzsl}}$ & $\boldsymbol{U_{gzsl}}$ & $\boldsymbol{H}$  \\
% P1-study on caddy, 
% P2-2019_Diver_Gesture_Recognition_using_Deep_Learning
% P3-2023_IEEE_ELEXCOM_Underwater_Diver_Gesture_Recognition
\midrule
AlexNet~\cite{yang2019diver}	&82.89 &$71.75 \pm 2.19$	&$0.76 \pm 1.21$	&$1.48 \pm 2.35$ \\
VGG-16~\cite{yang2019diver}	&95.00 &$76.20 \pm 1.23$	&$0.42 \pm 0.57$	&$0.84 \pm 1.12$  \\
ResNet-18~\cite{mangalvedhekar2023underwater} &98.00 & $53.54 \pm 3.77$ & $0.72 \pm 1.25$ & $1.38 \pm 2.39$ \\
ResNet-50~\cite{martija2020underwater} &97.06 & $61.94 \pm 2.86$ & $0.79 \pm 1.27$ & $1.53 \pm 2.44$ \\
% ResNet-101~\cite{pengcaddyian} &90.41 & $65.02 \pm 0.69$ &$ 0.72 \pm 1.13$ & $1.44 \pm 2.24$ \\
GoogleNet~\cite{yang2019diver}	&90.08 &$53.19 \pm 3.46$ & $0.8 \pm 1.39$ & $1.53 \pm 2.64$ \\
MobileNet-v3~\cite{mangalvedhekar2023underwater} &84.32 & $62.78 \pm 5.21$ & $1.31 \pm 2.23$ & $2.45 \pm 4.18$ \\
\bottomrule
\end{tabular}
\end{table}

% \begin{table}[t]
% \centering
% \caption{Method Comparison}
% \label{tab:my_label}
% % \begin{tabular}{lcccc}
% \begin{tabular}{l c c c c@{\hspace{10pt}} c@{\hspace{10pt}} c@{\hspace{10pt}} }

% \toprule
% Method &sup(P1) & sup(P2) & sup(P3) & Seen & Unseen & HM  \\
% % P1-study on caddy, 
% % P2-2019_Diver_Gesture_Recognition_using_Deep_Learning
% % P3-2023_IEEE_ELEXCOM_Underwater_Diver_Gesture_Recognition
% \midrule
% ResNet-18 &95.17
% & 88.59 &98 & $53.54 \pm 3.77$ & $0.72 \pm 1.25$ & $1.38 \pm 2.39$ \\
% ResNet-50 &94.91&- & 92.3 & $61.94 \pm 2.86$ & $0.79 \pm 1.27$ & $1.53 \pm 2.44$ \\
% ResNet-101 &90.41 &- &- & $65.02 \pm 0.69$ &$ 0.72 \pm 1.13$ & $1.44 \pm 2.24$ \\
% MobileNet-v3 &-&- & 84.32 & $62.78 \pm 5.21$ & $1.31 \pm 2.23$ & $2.45 \pm 4.18$ \\
% Googlenet	&-& 90.08 &- &$53.19 \pm 3.46$ & $0.8 \pm 1.39$ & $1.53 \pm 2.64$ \\
% Alexnet	&-& 82.89 &- &$71.75 \pm 2.19$	&$0.76 \pm 1.21$	&$1.48 \pm 2.35$ \\
% Vgg16	&-& 95.0 &- &$76.20 \pm 1.23$	&$0.42 \pm 0.57$	&$0.84 \pm 1.12$  \\
% \bottomrule
% \end{tabular}
% \end{table}

\subsubsection{Impact of activation functions}
In Tab.~\ref{tab:actFn}, we study the choice of activation function in the architecture of our novel gesture decoder while training the GCAT. In the CZSL scenario, we see comparable performance of the five different activation functions we studied. Interestingly, in the GZSL setting, we find that different activations have varying impacts on recognizing seen and unseen gestures. For example, the sigmoid function generalizes well, but seen accuracy is low. On the other hand, using SiLU~\cite{elfwing2018sigmoid} degrades unseen accuracy. The activation we used finally, GELU, performs well in both CZSL and GZSL settings. In summary, it is evident that the activation functions can dictate the quality of visual gesture features which can, in turn, hamper a generative method's ability to generate discriminative visual features for seen and unseen classes.

\section{Conclusion}
\label{sec:conclusion}
We introduced the task of zero-shot underwater gesture recognition in this paper and discussed its potential for diver-AUV communications. We proposed a two-stage framework consisting of a novel transformer that learns strong visual gesture representations and a conditional generative network that learns the distributions of these features. From the dataset perspective, we proposed three random seen-unseen splits of the CADDY dataset and reported our results in both conventional and generalized zero-shot settings, comparing them with state-of-the-art classification methods. Through extensive experimentations and ablations, we discussed how poorly supervised methods perform in zero-shot settings, why existing zero-shot classification methods suffer so much from class imbalance and bias problems, and how activation functions may impact transformers for gesture recognition. One of the areas for future research could be learning the semantic representations of gestures instead of using handcrafted prompt templates. We hope that this work can serve as a benchmark for the advancement of ZSUGR as a field since there is a huge room for improvement.

\bibliographystyle{splncs04}
\bibliography{main}




% LNCS Homepage, \url{http://www.springer.com/lncs}. Last accessed 4
% \end{thebibliography}
\end{document}